\documentclass{article}

\usepackage[preprint,nonatbib]{neurips_2018}

\usepackage{graphicx}
\usepackage{enumitem}   

\usepackage[utf8]{inputenc} % allow utf-8 input
\usepackage[T1]{fontenc}    % use 8-bit T1 fonts
\usepackage{hyperref}       % hyperlinks
\usepackage{url}            % simple URL typesetting
\usepackage{booktabs}       % professional-quality tables
\usepackage{amsfonts}       % blackboard math symbols
\usepackage{amsmath}
\usepackage{amssymb}
\usepackage{nicefrac}       % compact symbols for 1/2, etc.
\usepackage{microtype}      % microtypography
\usepackage{lipsum}
\usepackage{multicol}

\title{$k$-Nearest Neighbour Classifiers \\ $2^{nd}$ Edition (with Python examples)}

\author{
  P\'{a}draig Cunningham \\
  School of Computer Science\\
  University College Dublin\\
  \texttt{padraig.cunningham@ucd.ie} \\
   \And
 Sarah Jane Delany \\
  School of Computing\\
  Technological University Dublin\\
  \texttt{sarahjane.delany@tudublin.ie} \\
 }

\begin{document}
\maketitle

\begin{abstract}
Perhaps the most straightforward classifier in the arsenal or machine learning techniques is the Nearest Neighbour Classifier – classification is achieved by identifying the nearest neighbours to a query example and using those neighbours to determine the class of the query. This approach to classification is of particular importance because issues of poor run-time performance is not such a problem these days with the computational power that is available. This paper presents an overview of techniques for Nearest Neighbour classification focusing on; mechanisms for assessing similarity (distance), computational issues in identifying nearest neighbours and mechanisms for reducing the dimension of the data.

This paper is the second edition of a paper previously published as a technical report \cite{cunningham2007k}.
Sections on similarity measures for time-series, retrieval speed-up and intrinsic dimensionality have been added. An Appendix is included providing access to Python code for the key methods. 
\end{abstract}

% keywords can be removed
%\keywords{Nearest Neighbour \and Machine Learning \and Lazy Learning}

\section{Introduction}
The intuition underlying Nearest Neighbour Classification is quite straightforward, examples are classified based on the class of their nearest neighbours. It is often useful to take more than one neighbour into account so the technique is more commonly referred to as $k$-Nearest Neighbour ($k$-NN) Classification where $k$ nearest neighbours are used in determining the class. Since the training examples are needed at run-time, i.e. they need to be in memory at run-time, it is sometimes also called Memory-Based Classification. Because induction is delayed to run time, it is considered a Lazy Learning technique. Because classification is based directly on the training examples it is also called Example-Based Classification or Case-Based Classification.

\begin{figure}[t]
\centering
\includegraphics[width=6cm,height=5cm]{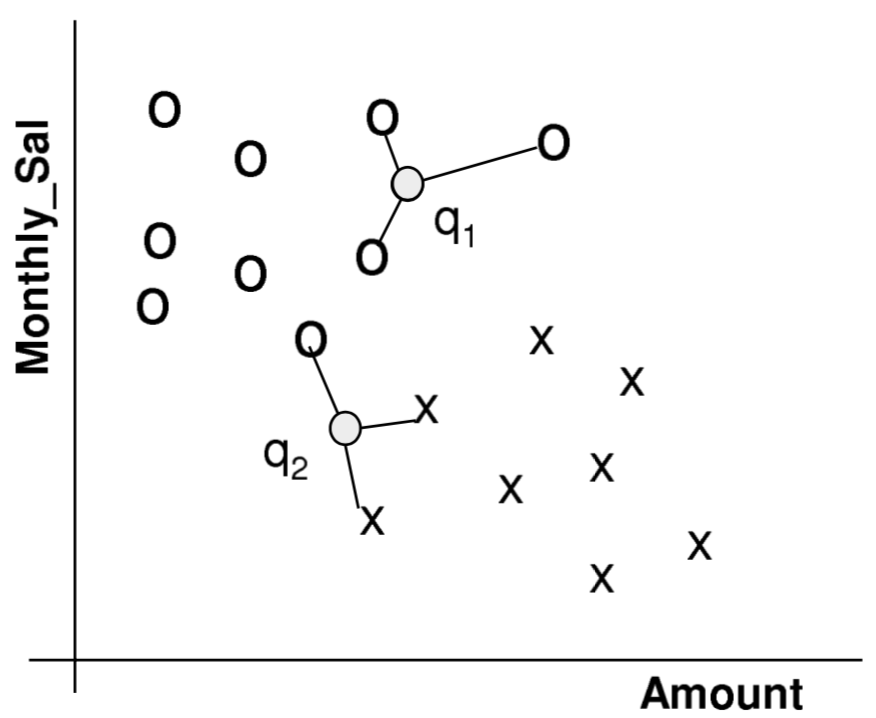}
\caption{3-Nearest Neighbour Classification in a 2D  feature space (Monthly$\_$Sal and Amount).}
\label{fig:3NN}
\end{figure}

The basic idea is as shown in Figure \ref{fig:3NN} which depicts a 3-Nearest Neighbour Classifier on a two-class problem in a two-dimensional feature space. In this example the decision for $q_1$ is straightforward – all three of its nearest neighbours are of class $O$ so it is classified as an $O$. The situation for $q_2$ is a bit more complicated at it has two neighbours of class $X$ and one of class $O$. This can be resolved by simple majority voting or by distance weighted voting (see below).
So $k$-NN classification has two stages; the first is the determination of the nearest neighbours and the second is the determination of the class using those neighbours.

Let us assume that we have a training dataset $D$ made up of $(\mathbf{x}_i)_{i \in [1,n]}$ 
training samples (where $ n = |D|$). The examples are described by a set of features $F$ and any numeric features have been normalised to the range [0,1]. Each training example is labelled with a class label $y_j \in Y$. Our objective is to classify an unknown example $\mathbf{q}$. For each $x_i \in D$ we can calculate the distance between $\mathbf{q}$ and $\mathbf{x}_i$ as follows:
\begin{equation}
d(\mathbf{q,x}_i)=\sum_{f \in F} w_f\delta(\mathbf{q}_f,\mathbf{x}_{if})
\label{eqn:dist}
\end{equation}
This is a summation over all the features in $F$ with $w_f$ the weight for each feature. There are a large range of possibilities for this distance metric; a basic version for continuous and discrete attributes would be:
\begin{equation}
\delta(\mathbf{q}_f,\mathbf{x}_{if}) = 
\begin{cases}
  0 & f \text{ discrete and } \mathbf{q}_f = \mathbf{x}_{if} \\    
  1 & f \text{ discrete and } \mathbf{q}_f \ne \mathbf{x}_{if} \\    
  |\mathbf{q}_f - \mathbf{x}_{if}| & f \text{ continuous}\\    
\end{cases}
\label{eqn:dist_feat}
\end{equation}

The $k$ nearest neighbours are selected based on this distance metric. Then there are a variety of ways in which the $k$ nearest neighbours can be used to determine the class of $\mathbf{q}$. The most straightforward approach is to assign the majority class among the nearest neighbours to the query.

It will often make sense to assign more weight to the nearer neighbours in deciding the class of the query. A fairly general technique to achieve this is distance weighted voting where the neighbours get to vote on the class of the query case with votes weighted by the inverse of their distance to the query.

\begin{equation}
Vote(y_j)=\sum_{c=1}^k \frac{1}{d(\mathbf{q,x}_c)^p}1(y_j,y_c)
\label{eqn:vote}
\end{equation}
Thus the vote assigned to class $y_j$ by neighbour $x_c$ is 1 divided by the distance to that neighbour, i.e. $1(y_j , y_c)$ returns 1 if the class labels match and 0 otherwise. In equation \ref{eqn:vote} $p$ would normally be 1 but values greater than 1 can be used to further reduce the influence of more distant neighbours.

Another approach to voting is based on Shepard’s work \cite{shepard1988toward} and uses an exponential function rather than inverse distance, i.e:

\begin{equation}
Vote(y_j)=\sum_{c=1}^k e^{-d(\mathbf{q,x}_c)}1(y_j,y_c)
\label{eqn:vote_exp}
\end{equation}

It is worth mentioning that $k$-NN can also be effective for regression \cite{altman1992introduction}. In regression the dependant variable $y$ is a real number ($y \in \mathbf{R}$) so the predicted value $\hat{y}$ can be the mean or weighted mean of the $y$ value for the neighbours. The weighted mean would be defined as follows:

\begin{equation}
\hat{y}=\frac{1}{k}\sum_{c=1}^k \frac{1}{d(\mathbf{q,x}_c)^p}y_c
\label{eqn:regression}
\end{equation} 
In this paper we consider three important issues that arise with the use of $k$-NN. In the next section we look at the core issue of similarity and distance measures and explore some \emph{exotic} (dis)similarity measures to illustrate the generality of the $k$-NN idea. In section 3 we look at computational complexity issues and review some speed-up techniques for $k$-NN. In section 4 we look at dimension reduction – both feature selection and sample selection. Dimension reduction is of particular importance with $k$-NN as it has a big impact on computational performance and accuracy. The paper concludes with a summary of the advantages and disadvantages of $k$-NN.

\section{Similarity and Distance Metrics}
\label{sec:metrics}
While the terms \emph{similarity metric} and \emph{distance metric} are often used colloquially to refer to any measure of affinity between two objects, the term metric has a formal meaning in mathematics. A metric must conform to the following four criteria (where $d(x,y)$ refers to the distance between two objects $x$ and $y$):

\vbox{% prevent page break in list
\begin{enumerate}
\item $d(x, y) \ge 0$; non-negativity
\item $d(x,y) = 0$ only if $x = y$; identity
\item $d(x, y)$ = $d(y, x)$; symmetry
\item $d(x, z) \ge d(x, y) + d(y, z)$; triangle inequality
\end{enumerate}}

It is possible to build a $k$-NN classifier that incorporates an affinity measure that is not a proper metric, however there are some performance optimisations to the basic $k$-NN algorithm that require the use of a proper metric \cite{schaaf1996fish,beygelzimer2006cover,zhu1999remembering,zhu1999remembering}. In brief, these techniques can identify the nearest neighbour of an object without comparing that object to every other object but the affinity measure must be a metric, in particular it must satisfy the triangle inequality.

The basic distance metric described in equations \ref{eqn:dist} and \ref{eqn:dist_feat} is a special case of the Minkowski Distance metric – in fact it is the 1-norm ($L_1$)Minkowski distance. The general formula for the Minkowski distance is
\begin{equation}
MD_p(\mathbf{q,x}_i)=\left( \sum_{f \in F} |\mathbf{q}_f - \mathbf{x}_{if}|^p\right)^\frac{1}{p}
\label{eqn:minkowski}
\end{equation}

The $L_1$ Minkowski distance is the Manhattan distance and the $L_2$ distance is the Euclidean distance. It is unusual but not unheard of to use $p$ values greater than 2. Larger values of $p$ have the effect of giving greater weight to the attributes on which the objects differ most. To illustrate this we can consider three points in 2D space; $A = (1,1),B = (5,1)$ and $C = (4,4)$. Since $A$ and $B$ differ on one attribute only the $MD_p(A,B)$ is 4 for all $p$, whereas $MD_p(A,C)$ is 6, 4.24 and 3.78 for $p$ values of 1, 2 and 3 respectively. So $C$ becomes the nearer neighbour to $A$ for $p$ values of 3 and greater.

The other important Minkowski distance is the $L^\infty$ or Chebyshev distance. 

\begin{equation}
MD_\infty(\mathbf{q,x}_i)=\max_{f \in F} |\mathbf{q}_f - \mathbf{x}_{if}|
\notag
\end{equation}

This is simply the distance in the dimension in which the two examples are most different; it is sometimes referred to as the chessboard distance as it is the number of moves it takes a chess king to reach any square on the board.

While the Euclidean and Manhattan distances are probably the most popular $k$-NN distance measures, much of the usefulness of $k$-NN derives from the potential to work with metrics that are specific to the data under analysis. In the next subsections we will look at the merits of Cosine Similarity and (Pearson) Correlation. Then we will look at some more complex distance measures that are specialised for particular data types, i.e. Earth Mover Distance for image data and Dynamic Time Warping for time-series data. 

\subsection{Cosine Similarity}\label{sec:cos}
\begin{figure}[h]
\centering
\includegraphics[width=6cm]{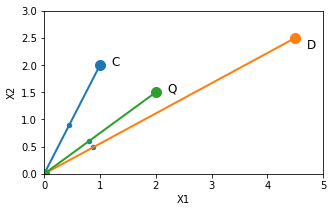}
\caption{Cosine Similarity: Q, C \& D are three examples in a 2D feature space, using Cosine Similarity the nearest neighbour to Q is D. }
\label{fig:cos}
\end{figure}

Like Minkowski distance, Cosine Similarity works with feature vector data. However, similarity is based on the angles between the feature vectors -- see Figure \ref{fig:cos}. While C would be the closer example to Q based on Euclidean distance, D is closer to Q when the angles between the features vectors is considered. 
The Cosine similarity between a query $\mathbf{q}$ and $\mathbf{x}_i$ is as follows:

\begin{equation}
    \text{Cos}(\mathbf{q,x}_i)= 
    \frac{\sum_{f \in F}\mathbf{q}_f \cdot \mathbf{x}_{if}}
    {\sqrt{\sum_{f\in F} \mathbf{q}_f^2} \sqrt{\sum_{f\in F} \mathbf{x}_{if}^2}}
\end{equation}\label{eq:cos}

This is the dot product of the feature values normalised by the lengths of the feature vectors. Cosine similarity is a popular metric in text analytics. When  text is processed as a bag of words the features are word counts and Cosine similarity has the advantage that it is independent of the magnitude of the feature vectors. Thus it is insensitive to document size. Cosine similarity requires that all feature values are positive real numbers. 

If feature values are positive then the Cosine similarity will be in the interval $[0,1]$. So we can define a Cosine distance measure: 
\begin{equation}
    \text{CosD}(\mathbf{q,x}_i) = 1 -  \text{Cos}(\mathbf{q,x}_i)
    \label{eqn:CosD}
\end{equation}

\subsection{Correlation}\label{sec:corr}
\begin{figure}[h]
\centering
\includegraphics[width=6cm]{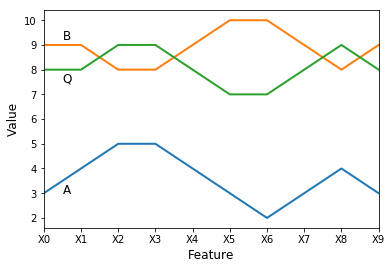}
\caption{Correlation: Q, A \& B are three examples described by 10 features (X0 - X9), correlation recognises that Q is more similar to A than to B.}
\label{fig:correl}
\end{figure}

Figure \ref{fig:correl} shows a scenario where correlation would be the  appropriate similarity measure. While B is the more similar example to the query Q in terms of feature values, the \emph{pattern} in A  better correlates with  Q. Sometimes this correlation is is the key to the underlying similarity. This would be appropriate where the feature values reflect resource allocation (for example household expenditure on 10 categories (X0 - X9)). The magnitudes might be quite different but the allocation pattern could be the same. 

The two most popular correlation coefficients are the Pearson and Spearman measures \cite{de2016comparing}. 
The Pearson is applicable for features that are normally distributed. When reference is made to a correlation coefficient without specifying which one, it is probably the Pearson. 
The Pearson correlation between a query $\mathbf{q}$ and a sample $\mathbf{x}_i$ is defined as follows:
\begin{equation}
    r(\mathbf{q,x}_i)= \frac{\sum_{f\in F}(\mathbf{q}_f-\Bar{\mathbf{q}})
    (\mathbf{x}_{if}-\Bar{\mathbf{x}}_i)}
    {(n-1)s_qs_{x_i}}
    =\frac{\sum_{f\in F}(\mathbf{q}_f-\Bar{\mathbf{q}})
    (\mathbf{x}_{if}-\Bar{\mathbf{x}}_i)}
    {\sqrt{\sum_{f\in F}(\mathbf{q}_f-\Bar{\mathbf{q}})^2
    \sum_{f\in F}(\mathbf{x}_{if}-\Bar{\mathbf{x}}_i)^2
    }}
    \end{equation}
where $\Bar{\mathbf{x}}_i$ and $s_x$ are the mean and standard deviation of $\mathbf{x}_i$. This is the dot product of the mean-adjusted $\mathbf{q}$ and $\mathbf{x}_i$ vectors divided by their standard deviations. This mean adjustment makes the measure insensitive to variations in scale. 

In circumstances where the features are not normally distributed the Spearman (rank) correlation can be used. The feature values are ranked and the statistic is calculated using ranks rather than the original values. 

Correlation scores range from $[-1,1]$. A score of 1 represents a perfect correlation, 0 is no correlation and -1 means the samples are anti-correlated. A correlation is a similarity score and so can be converted to a distance in the same manner as for Cosine -- see equation \ref{eqn:CosD}.

\subsection{Other Distances Metrics for Multimedia Data}
The Minkowski distance defined in (\ref{eqn:minkowski}) is a very general metric that can be used in a $k$-NN classifier for any data that is represented as a feature vector. When working with image data a convenient representation for the purpose of calculating distances is a colour histogram. An image can be considered as a grey-scale histogram $H$ of $N$ levels or bins where $h_i$ is the number of pixels that fall into the interval represented by bin $i$ (this vector $h$ is the feature vector). The Minkowski distance formula (\ref{eqn:minkowski}) can be used to compare two images described as histograms. $L_1$, $L_2$ and less often $L_\infty$ norms are used.
Other popular measures for comparing histograms are the Kullback-Leibler divergence (\ref{eqn:KL}) \cite{kullback1951information} and the $\chi^2$ statistic (\ref{eqn:chi}) \cite{rubner2000earth}.

\begin{equation}
d_{KL}(H,K)=\sum_{i=1}^{N} h_i \text{ log}\left(\frac{h_i}{k_i}\right)
\label{eqn:KL}
\end{equation}

\begin{equation}
d_{\chi^2}(H,K)=\sum_{i=1}^{N} \frac{h_i- m_i}{h_i}
\label{eqn:chi}
\end{equation}
where $H$ and $K$ are two histograms, $h$ and $k$ are the corresponding vectors of bin values and $m_i = \frac{h_i+k_i}{2}$.

While these measures have sound theoretical support in information theory and in statistics they have some significant drawbacks. The first drawback is that they are not metrics in that they do not satisfy the symmetry requirement. However, this problem can easily be overcome by defining a modified distance between $x$ and $y$ that is in some way an average of $d(x, y)$ and $d(y, x)$ – see \cite{rubner2000earth} for the Jeffrey divergence which is a symmetric version of the Kullback-Leibler divergence.

A more significant drawback is that these measures are prone to errors due to bin boundaries. The distance between an image and a slightly darker version of itself can be great if pixels fall into an adjacent bin as there is no consideration of adjacency of bins in these measures.

\textbf{Earth Mover Distance} The Earth Mover Distance (EMD) is a distance measure that overcomes many of these problems that arise from the arbitrariness of binning. As the name implies, the distance is based on the notion of the amount of effort required to convert one image to another based on the analogy of transporting \emph{mass} from one distribution to another. If we think of two images as distributions and view one distribution as a mass of earth in space and the other distribution as a hole (or set of holes) in the same space then the EMD is the minimum amount of work involved in filling the holes with the earth.

In their analysis of the EMD Rubner \emph{et al.} argue that a measure based on the notion of a \emph{signature} is better than one based on a histogram. A signature $\{\mathbf{s}_j = \mathbf{m}_j , w_{\mathbf{m}_j} \}$ is a set of $j$ clusters where $\mathbf{m}_j$ is a vector describing the mode of cluster $j$ and $w_{\mathbf{m}_j}$ is the fraction of pixels falling into that cluster. Thus a signature is a generalisation of the notion of a histogram where boundaries and the number of partitions are not set in advance; instead $j$ should be ‘appropriate’ to the complexity of the image \cite{rubner2000earth}.

\begin{figure}[t]
\centering
\includegraphics[width=6cm,height=5cm]{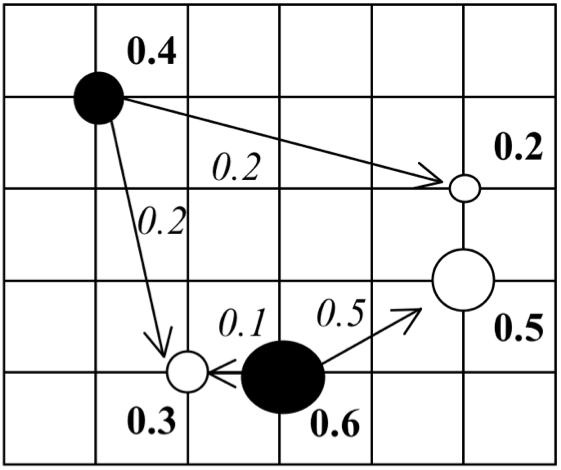}
\caption{An example of the EMD between two 2D signatures with two points (clusters) in one signature and three in the other (based on example in \cite{rubner1997earth}).}
\label{fig:EMD}
\end{figure}

The example in Figure \ref{fig:EMD} illustrates this idea. We can think of the clustering as a quantization of the image in some colour space so that the image is represented by a set of cluster modes and their weights. In the figure the source image is represented in a 2D space as two points of weights 0.6 and 0.4; the target image is represented by three points with weights 0.5, 0.3 and 0.2. In this example the EMD is calculated to be the sum of the amounts moved (0.2, 0.2, 0.1 and 0.5) multiplied by the distances they are moved. Calculating the EMD involves discovering an assignment that minimizes this amount.

For two images described by signatures $S =\{\mathbf{m}_j,w_{\mathbf{m}_j}\}^n_{j=1}$ and $Q=\{\mathbf{p}_k,w_{\mathbf{p}_k}\}^r_{k=1}$ we are interested in the work required to transfer from one to the other for a given flow pattern $\mathbf{F}$:
\begin{equation}
    WORK(S,Q,\mathbf{F})=\sum_{j=1}^n\sum_{k=1}^r d_{jk}f_{jk}
\end{equation}

where $d_{jk}$ is the distance between clusters $\mathbf{m}_j$ and $\mathbf{p}_k$ and $f_{jk}$ is the flow between $\mathbf{m}_j$ and $\mathbf{p}_k$ that minimises overall cost. An example of this in a 2D colour space is shown in Figure \ref{fig:EMD}. Once the transportation problem of identifying the flow that minimises effort is solved (using dynamic programming) the EMD is defined to be:
\begin{equation}
    EMD(S,Q)=\frac{\sum_{j=1}^n\sum_{k=1}^r d_{jk}f_{jk}}{\sum_{j=1}^n\sum_{k=1}^r f_{jk}}
\end{equation}

Efficient algorithms for the EMD are described in \cite{rubner2000earth} however this measure is expensive to compute with cost increasing more than linearly with the number of clusters. Nevertheless it is an effective measure for capturing similarity between images.

\textbf{Compression-Based Dissimilarity} In recent years the idea of basing a similarity metric on compression has received a lot of attention. \cite{li2004similarity,keogh2004towards}. Indeed Li \emph{et al.} \cite{li2004similarity}, refer to this as \emph{The} similarity metric. The basic idea is quite straight-forward; if two documents are very similar then the compressed size of the two documents concatenated together will not be much greater than the compressed size of a single document. This will not be true for two documents that are very different. Slightly more formally, the difference between two documents $A$ and $B$ is related to the compressed size of document $B$ when compressed using the codebook produced when compressing document $A$.

The theoretical basis of this metric is in the field of Kolmogorov complexity, specifically in conditional Kolmogorov complexity.
\begin{equation}
    d_{Kv}(x,y)=\frac{Kv(x|y)+Kv(y|x)}{Kv(xy)}
\end{equation}
where $Kv(x|y)$ is the length of the shortest program that computes $x$ when $y$ is given as an auxiliary input to the program and $Kv(xy)$ is the length of the shortest program that outputs $y$ concatenated to $x$. While this is an abstract idea it can be approximated using compression:
\begin{equation}
    d_{C}(x,y)=\frac{C(x|y)+C(y|x)}{C(xy)}
\end{equation}
$C(x)$ is the size of data $x$ after compression, and $C(x|y)$ is the size of $x$ after compressing it with the compression model built for $y$. If we assume that $Kv(x|y) = Kv(xy) - Kv(y)$ then we can define a normalised compression distance:
\begin{equation}
    d_{NC}(x,y)=\frac{C(xy)-\text{min}(C(x),C(y))}{\text{min}(C(x),C(y))}
\end{equation}
It is important that $C(.)$ should be an appropriate compression metric for the data. Delany and Bridge \cite{delany2007} show that compression using Lempel-Ziv (GZip) is effective for text. They show that this compression based metric is more accurate in $k$-NN classification than distance based metrics on a bag-of-words representation of the text.

\subsection{Similarity Metrics for Time-Series}\label{sec:DTW}
\begin{figure}[t]
\begin{multicols}{2}
    \includegraphics[width=\linewidth]{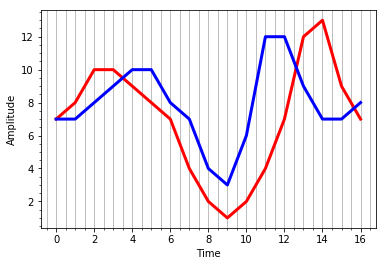}
     \par 
    \includegraphics[width=0.75\linewidth]{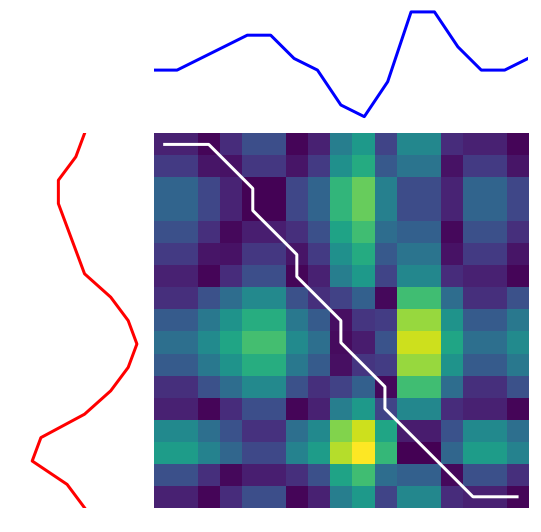}\par 
\end{multicols}
\caption{The image on the left illustrates the main challenge in quantifying similarity between time-series. The two series are similar but the Euclidean distance between them is large. The image on the right shows the DTW mapping between the two time-series (produced using \textsf{tslearn}\cite{tavenard2017tslearn}).}
\label{fig:DTW}%
\end{figure}

Time-series data can be as diverse as human activity   measured by wearable sensors \cite{mahato2019scoring} or measurements coming from a manufacturing process. 
There is a long history of Machine Learning research on time-series analysis and 
 $1$-NN is the base-line metric for time-series classification \cite{bagnall2017great}. However, the special characteristics of time-series do present challenges for $k$-NN. Consider a query time-series $\mathbf{q}$ and a target $\mathbf{x}$:
\begin{equation}
    \mathbf{q}=q_1,q_2,...,q_j,...,q_m
 \end{equation}
\begin{equation}
    \mathbf{x}=x_1,x_2,...,x_j,...,x_n
\end{equation}
 While both time-series are vectors, the Euclidean distance between these two vectors may be quite large even if they have the same general shape (see Figure \ref{fig:DTW}). Furthermore the two time-series might be of different lengths. To complicate things further, similarity might depend on specific features (motifs) in the time-series rather than similarity across the time-series as a whole. 

A number of methods for scoring similarity between time-series have been developed that allow $k$-NN to work with time-series data. Three popular methods are:
\begin{itemize}
    \item \textbf{Dynamic Time Warping (DTW):} Because two time series may be fundamentally similar but offset or slightly distorted, DTW allows the time axis to be \emph{warped} to identify underlying similarities \cite{keogh2001derivative}.
    \item \textbf{Symbolic Aggregate Approximation (SAX):} The idea with SAX is to discretize the time series so that it can be represented as a sequence of symbols \cite{lin_keogh_lonardi_chiu_2003}. Then methods for scoring sequence similarity can be applied. 
    
    \item \textbf{Symbolic Fourier Approximation (SFA):} SFA is like SAX except the sequence representation is produced from a discrete Fourier transform representation of the signal rather than a discretization of the signal itself. So SFA is a frequency domain rather than a time domain representation of the signal \cite{schafer_hogqvist_2012}. 
\end{itemize}

By far the most popular of these is DTW so we will provide some detail here on how DTW works. As the name suggests, the idea is to allow the time-series to be stretched (warped) to find the best mapping. The DTW distance is defined as follows:

\begin{equation}
    DTW(\mathbf{q,x})=\underset{\pi}{\text{min}}\sqrt{\sum_{(i,j)\in\pi}{d(q_i,x_j)^2}}
\end{equation}

where $\pi=[\pi_1,...,\pi_l,...,\pi_L]$ is the optimum path (mapping) having the following properties:
\begin{itemize}
    \item $m = |\mathbf{q}|,n = |\mathbf{x}|$
    \item $\pi_1 =(1,1), \pi_L = (m,n)$
    \item $\pi_{l+1} - \pi_l \in\{(1,0),(0,1),(1,1)\}$
\end{itemize}

The DTW path for the two time-series in Figure \ref{fig:DTW} is shown on the right. It starts at the top left (1,1) and finishes at the bottom right ($m,n$). Each point $(i,j)$ on the path indicates the mapping between $q_i$ and $x_j$.
The extent of the deviation from the main diagonal reflects the \emph{warping}. In practice, the path may be restricted to a band around the main diagonal to restrict warping. The computational complexity of DTW is $O(n,m)$ because it entails a search through the matrix shown on the right is Figure \ref{fig:DTW}. This is effectively $O(n^2)$ in the length of the time-series -- so DTW is computationally expensive. 

Finally, DTW is not a proper metric because it fails two of the criteria laid out at the beginning of this section. $DTW(\mathbf{q,x}) = 0 \nRightarrow \mathbf{x} = \mathbf{x}$ and the triangle inequality may not hold. This means that speed-up mechanisms such as Ball Trees (section \ref{sec:BallT}) that work for proper similarity metrics cannot be applied. Neither can mechanisms that work for vector space representations, i.e. Kd-Trees (section \ref{sec:kdtree}) and Random Projection Trees (section \ref{sec:RPTree}).
\section{Computational Complexity}\label{sec:Complexity}
Computationally expensive metrics such as the Earth-Mover’s Distance and compression based (dis)similarity metrics focus attention on the computational issues associated with k-NN classifiers. Basic $k$-NN classifiers that use a simple Minkowski distance will have a time behaviour that is $O(dn)$ where $n$ is the number of data samples and $d$ is the number of features that describe the data, i.e. the distance metric is linear in the number of features and the comparison process increases linearly with the amount of data. The computational complexity of the EMD and compression metrics is more difficult to characterise but a $k$-NN classifier that incorporates an EMD metric is likely to be $O(nc^3 \text{log} c)$ where $c$ is the number of clusters \cite{rubner2000earth}.

For these reasons there has been considerable research on editing down the training data and on reducing the number of features used to describe the data (see section \ref{sec:DimRed}). There has also been considerable research on alternatives to the exhaustive search strategy (brute force) that is used in the standard $k$-NN algorithm. In the remainder of this section we review Kd-Trees and Cover Trees, the two speedup strategies included in Scikit-learn. We also review some approximate $k$-NN algorithms that don't guarantee to retrieve nearest neighbours but offer dramatic speedup with little loss of accuracy. In the final sub-section a simple comparison of Kd-Trees and Cover Trees against brute force search is presented. 

\subsection{Kd Trees}\label{sec:kdtree}
Kd-Trees represent the longest established strategy for speedup in $k$-NN \cite{bentley1975multidimensional}. It is best to think of Kd-Trees as a general strategy rather than a single algorithm. The general idea is that a binary tree is used to successively partition the dataset with training samples sorted to the leaves of the tree. This offers the potential for retrieval time that is $O(d\log(n))$ rather than $O(dn)$. 

\begin{figure}[t]
\centering
\includegraphics[height=5cm]{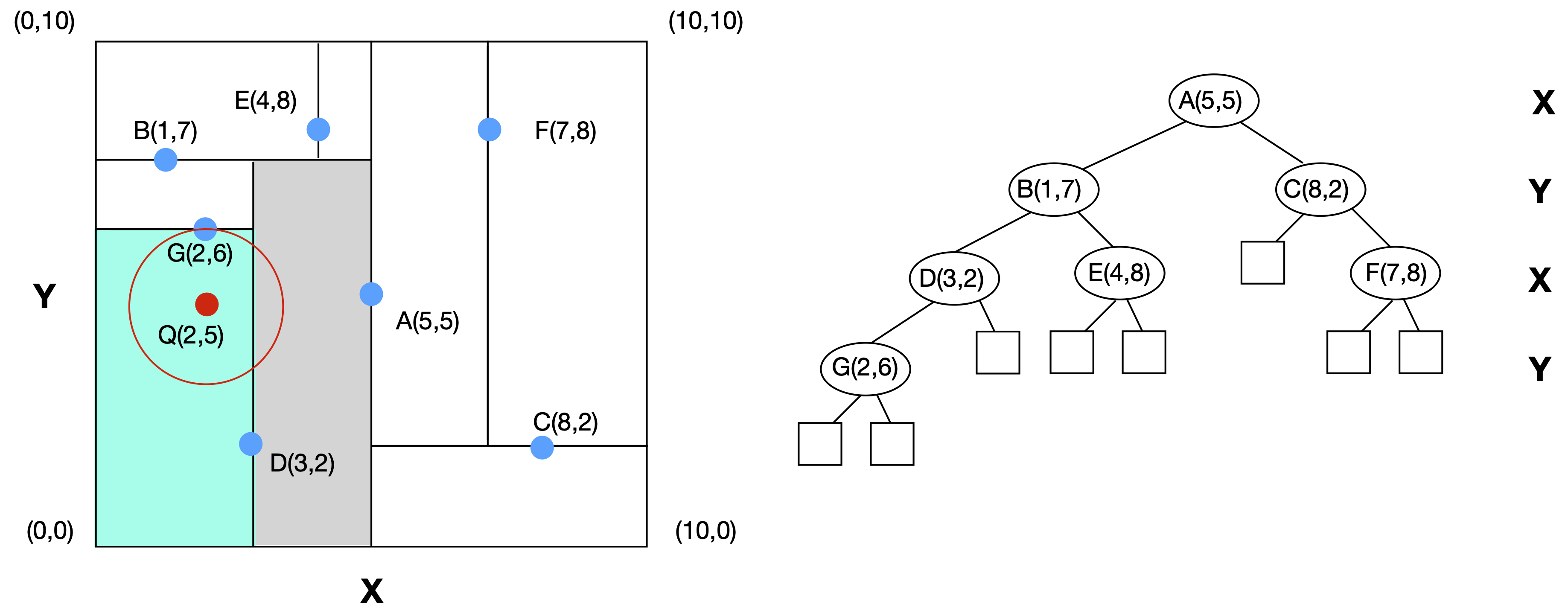}
\caption{A Kd-Tree based on the example in the original paper by Bentley \cite{bentley1979multidimensional}). The partitioning of the 2D feature space shown on the left corresponds to the tree on the right.}
\label{fig:kDT}
\end{figure}

A sample Kd-Tree is shown in Figure \ref{fig:kDT}. The data is described by two features so it can be represented as a 2D plot. The plot on the left corresponds to the binary tree on the right. The Kd-Tree always partitions the data along hyperplanes (lines in the 2D case) that are perpendicular to the axes. 

The figure shows a query point $Q(2,5)$. The search for nearest neighbours for $Q$ will locate it to the appropriate node in the tree $G(2,6)$. It can be seen in the plot that the nearest neighbour for $Q$ is not guaranteed to be located in the hypercube represented by $G$. However the distance between $G$ and $Q$ gives us an upper bound on the distance to the nearest neighbour. It is clear that the grey box (hypercube) needs also to be considered; the rest of the tree can be bounded out from consideration. It is this potential to bound out large parts of the data that yields the $O(d\log(n))$ performance. 

Some other aspects of Kd-Trees that need to be considered are as follows:
\begin{itemize}
    \item \textbf{Constructing Kd-Trees} entails a straightforward binary partitioning of the data and sorting the data to leaf nodes so the construction process is comparatively quick. The partition is typically at the median value for the selected feature.
    \item At each step in the building of the tree a decision has to be made on \textbf{feature selection}. The policy could be to cycle through the features in order or to select the feature in which the variance (spread) in the data is highest. 
    \item The query time will increase with the number of neighbours required ($k$). For very large $k$ query time will exceed that for brute force search. 
    \item The $O(d\log(n))$ retrieval time depends on a \textbf{balanced tree}. If the tree is not well balanced some retrieval times will be poor \cite{bentley1975multidimensional}.
    \item The main drawback with Kd-Trees is that the \textbf{curse of dimensionality} still applies. The benefits of using a binary tree as an indexing structure cease to apply when $d$ is large (say $>20$). Kleinberg \cite{kleinberg1997two} points out that when $d > log(n)$,  $O(d\log(n))$ is no better than $O(dn)$. 
\end{itemize}

\subsection{Ball Trees}\label{sec:BallT}
Figure \ref{fig:kDT} shows that a Kd-Tree indexes the data by partitioning the feature space. By contrast a Ball Tree is a `metric tree' in the sense that it is based on a metric defined on pairs of samples \cite{fukunaga1975branch,zhu1999remembering}. The construction of the ball tree is akin to a hierarchical clustering problem that can be tackled top-down or bottom-up:
\begin{itemize}
    \item \textbf{Bottom-up:} Initially each data point is a point sized bounding ball. At each step, select the closest pair of balls, the pair that have the smallest bounding ball that covers them. Join these balls. Continue until the top-level bounding ball is reached. 
   
    \item \textbf{Top-down:} At each step, two data points are chosen that have the maximum distance between them. The remaining points are partitioned by  allocating to the closer or these. This process is repeated recursively until a stopping criterion is met, e.g. number of samples at a leaf node.   
\end{itemize}

In contrast to Kd-Trees, the construction of a Ball tree depends on a metric defined on the data rather than a feature space representation. 
 However, it should be noted that the distance measure must be a metric so a Ball Tree cannot be applied for measures such as Earth Mover Distance or Dynamic Time Warping. Compared with Kd-Trees, Ball Trees have the potential to perform better for high-dimension data, for example in image analysis \cite{kumar2008good}.

\subsection{Approximate \textit{k}-NN}
Brute force search for $k$-NN is $O(dn)$. As we have seen in the preceeding sections we can get over the linear dependence on $n$ but high dimension data is still a problem. Fortunately, for many applications, it is not essential to retrieve the absolute nearest neighbours. For instance, in recommender systems, the most similar item is not necessarily required - indeed items that are reasonably close may offer some serendipitous discovery. In $k$-NN classification, neighbours that are close (but not necessarily closest) are probably of the correct class. So in this section we review the two most popular strategies for Approximate $k$-NN, these are Locality Sensitive Hashing and Random Projection Trees. 

\cite{liu2005investigation}

\subsubsection{Locality Sensitive Hashing} With Locality Sensitive Hashing (LSH) the objective is to map similar items into the same `buckets' with high probability. This contrasts with conventional hashing where the objective of minimising hashing collisions means that  similar items will have very different hashes. Given that LSH maps similar items to the same buckets it can be used to implement approximate nearest neighbour search. The strategy is to use a number of variants of LSH algorithms to retrieve a candidate set of nearest neighbours. This candidate set is the union of the items in the buckets returned by the LSH algorithms.  Then the similarity metric can be applied to these candidates to find nearest neighbours that will be near-optimal \cite{indyk1998approximate,liu2005investigation}.
\subsubsection{Random Projection Trees} \label{sec:RPTree}
In section \ref{sec:kdtree} we saw that exact nearest neighbour search is a two stage process. First the query is located to the correct leaf node in the tree and candidate nearest neighbours are identified. Then there is a backtracking process that finds better candidates or bounds out sections of the tree from consideration. The retrieval of nearest neighbours is guaranteed without explicitly  measuring against all data points. Random Projection Trees depends on two extensions to this basic kd-tree idea:
\begin{itemize}
    \item \textbf{Defeatist Search:} The query item is located to the correct leaf node but the backtracking process to ensure optimality is dropped or at least greatly curtailed \cite{liu2005investigation}. The search gives up early which might be considered a bit \textit{defeatist}. 
    \item \textbf{Multiple Trees:} If the search returns without backtracking the prospect of finding good neighbours can be improved by repeating with multiple trees. Different variants of the tree can be produced by including a random element in the Kd-Tree generation process \cite{silpa2008optimised}. Since there is a risk that there will not be great variety in the Kd-Tree variants, it is common to produce different trees by randomly projecting the data into a different space (i.e. perform a simple linear transformation on the data) \cite{kleinberg1997two}. 
\end{itemize}

In the next section we provide a demonstration of the effectiveness of Approximate $k$-NN on a number of datasets. The results show significant speed-up with little or no loss of accuracy. The caveat is that it only works for feature vector data.

\subsection{Speed-Up Evaluation}\label{sec:speedup}
The objective in this section is to show the potential speed-up that is possible with these methods, it is not meant as a comprehensive evaluation. In our first evaluation we assess the three options available in the $k$-NN implementation in scikit-learn (\url{scikit-learn.org)}. These are brute-force search, Kd-Trees and Ball Trees. The evaluation covers the four datasets summarised in Table \ref{tab:speedup_data}. Two of these datasets are low dimension ($<10$). The Credit dataset would be considered high-dimension with 23 features. 

\begin{table}[ht]
\caption{Overview of the datasets used to test the speedup algorithms.}
    \centering
\begin{tabular}{c|c c c c }
& HTRU & Shuttle & Letter & Credit \\ 
 \hline
Samples & 17,898 & 43,494 & 20,000 & 30,000 \\ 
Features & 8 & 9 & 16 &   23 \\
\hline
Winning Alg. & kd-tree & kd-tree & kd-tree & brute-force \\
\end{tabular}
\label{tab:speedup_data}
\end{table}
\begin{figure}[h]
\begin{multicols}{2}
    \includegraphics[width=\linewidth]{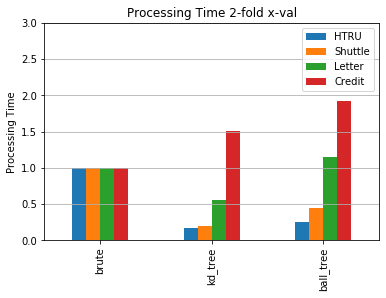}
     \par 
    \includegraphics[width=\linewidth]{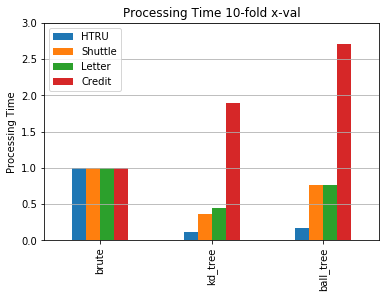}\par 
\end{multicols}
\caption{Processing time for Kd-Tree and Ball Tree compared with brute force search. Times are normalised to the brute force time. (< 1 is an improvement). 2-fold cross validation on the left and 10-fold on the right. }
\label{fig:kd+ball}%
\end{figure}
We present two sets of results (see Figure \ref{fig:kd+ball}), one using 2-fold cross validation and one using 10-fold. The objective is to show the impact of the tree building phase; while both cross validations use all the data for testing, the 10-fold cross validation incurs the tree building overhead 10 times instead of twice.

The bar charts show the processing time divided by the time for brute force search. It is clear that significant speed-up is possible for the low-dimension datasets. The Kd-Tree results are slightly better than Ball Tree in all cases.  However, the performance for the Credit dataset is worse than brute-force search. This is to be expected given that it is high-dimension.

\begin{figure}[ht]
\begin{multicols}{2}
    \includegraphics[width=\linewidth]{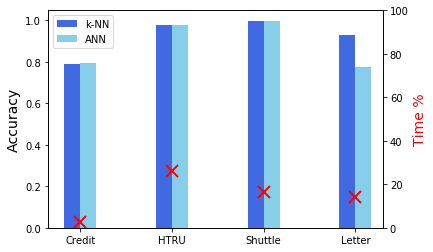}
     \par 
    \includegraphics[width=0.95\linewidth]{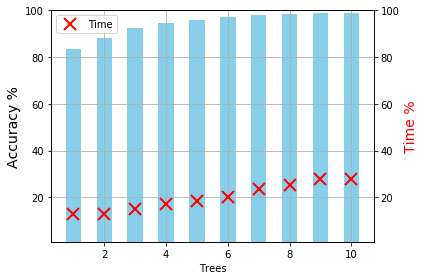}\par 
\end{multicols}
\caption{The plot on the left shows the accuracy of ANN using a single tree compared with `full search' $k$-NN. The time saving is significant. The plot on the right shows that accuracy on the Letter dataset can be improved with the addition of more trees.}
\label{fig:ANN}%
\end{figure}

This poor performance on high-dimension data shows that the curse of dimensionality cannot be avoided in exact nearest neighbour search. Figure \ref{fig:ANN} shows the speed-up that can be achieved with Approximate Nearest Neighbour. The method we evaluate is called Annoy (\url{github.com/spotify/annoy}) and uses Random Projection Trees \cite{Github:annoy}. Annoy stands for `Approximate Nearest Neighbor Oh Yeah' but the name probably stems from the fact that the method is annoying effective. The results in Figure \ref{fig:ANN} show dramatic speed-up with almost no loss of accuracy except for the Letter dataset. When more trees are added the accuracy reaches that achievable with exact $k$-NN with a four-fold improvement in processing time.

\section{Dimension Reduction} \label{sec:DimRed}

Given the high dimension nature of the data, Dimension Reduction is a core research topic in Machine Learning. Research on Dimension Reduction has itself two dimensions; the dimensions of a dataset of $|D|$ examples described by $|F|$ features can be reduced by selecting a subset of the examples or by selecting a subset of the features (an alternative to this is to transform the data into a representation with less features). It is important to emphasise that dimension reduction through feature selection can be achieved without loss of information because the \emph{intrinsic} dimension of the data may be considerably less than the number of features. This notion of intrinsic dimension is discussed in section \ref{sec:intD}.

Dimension reduction as achieved by supervised feature selection is described in section \ref{sec:FeatSel}. Unsupervised feature transformation using Principle Component Analysis (PCA) \cite{cunningham2008dimension} can be used as a preprocessing step for $k$-NN \cite{baek2002pca}. PCA is discussed in the next section in the context of intrinsic dimension.
However there is no evidence that PCA can be combined with $k$-NN    without sacrificing accuracy so PCA will not be covered in this paper. The other aspect of dimension reduction is the deletion of redundant or noisy instances in the training data -- this is reviewed in section \ref{sec:NoiseRed}.

\subsection{Intrinsic Dimension}\label{sec:intD}

\begin{figure}[h]
\centering
\includegraphics[width=0.7\linewidth]{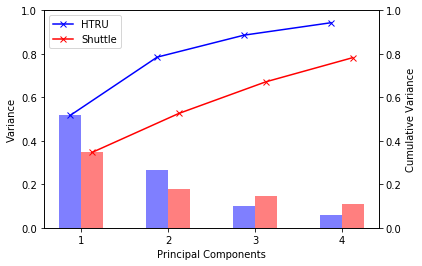}
\caption{The first four principal components of the HTRU and Shuttle datasets.}
\label{fig:PCA}
\end{figure}

Colloquially we can think of the intrinsic dimension as the minimum number of features required to provide a `good' representation of the data. This notion of a `good' representation can be considered in terms of Principle Component Analysis (PCA). We can represent a dataset $D$ as a rectangular matrix $\mathbf{D}$ of dimension $n\times p$, that is $n$ samples described by $p$ features. If we perform PCA on $\mathbf{D}$ we get:
\begin{equation}
    \mathbf{T}_{n \times r}=\mathbf{D}_{n\times p}\mathbf{W}_{p \times r}
\end{equation}
 The PCA provides a linear mapping of the data into a lower dimension representation $\mathbf{T}$. The PCA also provides a ranking of the principle components (PCs) in terms of the variance in the data that they capture. We can select the top $s$ PCs that together capture $(1-\epsilon)$ fraction  of the variance in the data. This $s$ is an approximation of the intrinsic dimension of the data.  
 \begin{equation}
    \mathbf{T}_{n \times s}=\mathbf{D}_{n\times p}\mathbf{W}_{p \times s}
\end{equation}
The variance captured by the first four PCs for the HTRU and Shuttle datasets is shown in Figure \ref{fig:PCA}. The four PCs capture almost all of the variance for the HTRU data but less than 80\% for Shuttle. We can think of four as a reasonable assessment of the intrinsic dimension of the HTRU data, whereas the intrinsic dimension for Shuttle is greater than four. 
 
This PCA inspired notion of intrinsic dimension is a global approximation and there may be parts of the space where the intrinsic dimension is locally less than $s$. Imagine a neighbourhood of radius $r$ around a point $\mathbf{q}$  (e.g. among the $k$ nearest neighbours),
$(1- \epsilon)$ fraction of the variance will be covered by $s^\prime$ features, where $s^\prime < p$. 

\begin{figure}[h]
\centering
\includegraphics[width=10cm]{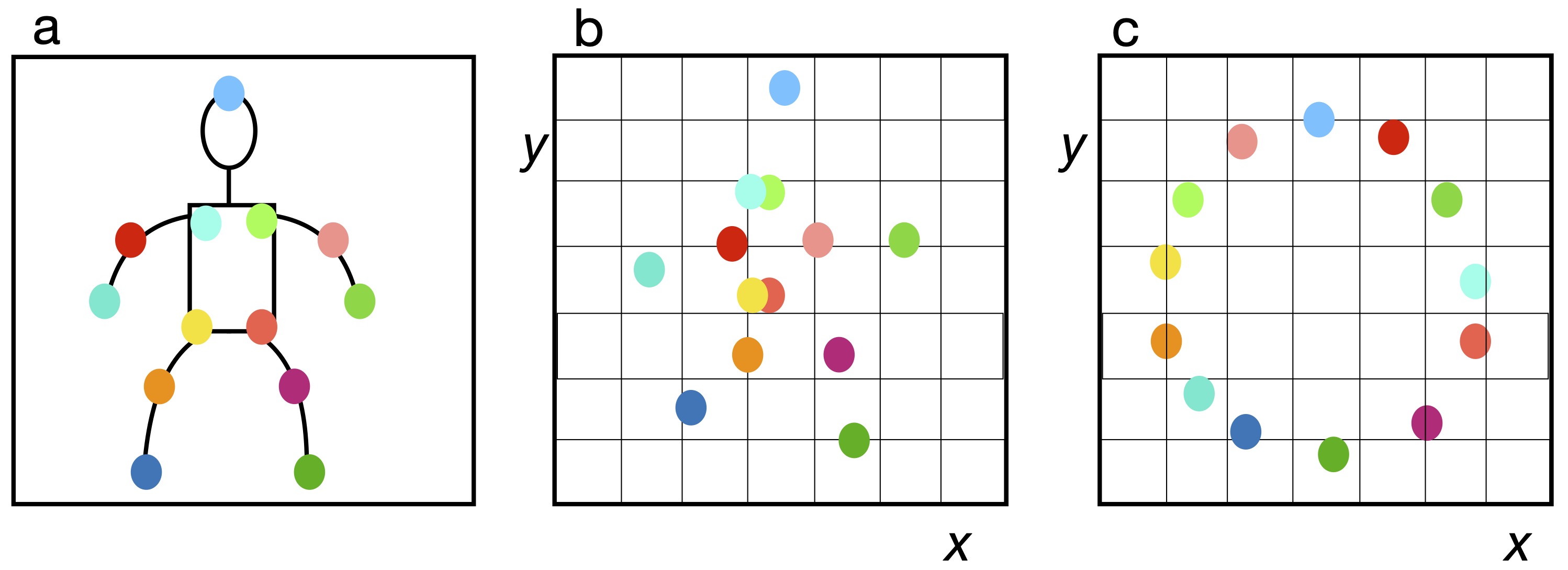}
\caption{Consider a simple motion capture system where 13 coloured balls capture the motion of the stick figure (a). (b) shows a valid configuration of these balls. (c) shows a configuration in this space that is not reachable. (Motivated by example in \cite{dasgupta2008random}.)}
\label{fig:ImpDim}
\end{figure}

Dasgupta \& Freund \cite{dasgupta2008random} provide an insightful example to explain intrinsic dimension. Imagine a motion capture system with 13 markers attached to a person to facilitate processing (see Figure \ref{fig:ImpDim}). In a 2D image these markers can be represented by 13 pairs of $x,y$ coordinates. So the dimension of the motion capture data will be 26. However, it is clear from Figure \ref{fig:ImpDim} (c) that very many points in the 26D space are not reachable. This system doesn't really have 26 degrees of freedom. Instead the person could be represented by the joint angles in the body, a number much smaller than 26.  

\subsection{Feature Selection}\label{sec:FeatSel}
\begin{figure}[t]
\centering
\includegraphics[width=10cm]{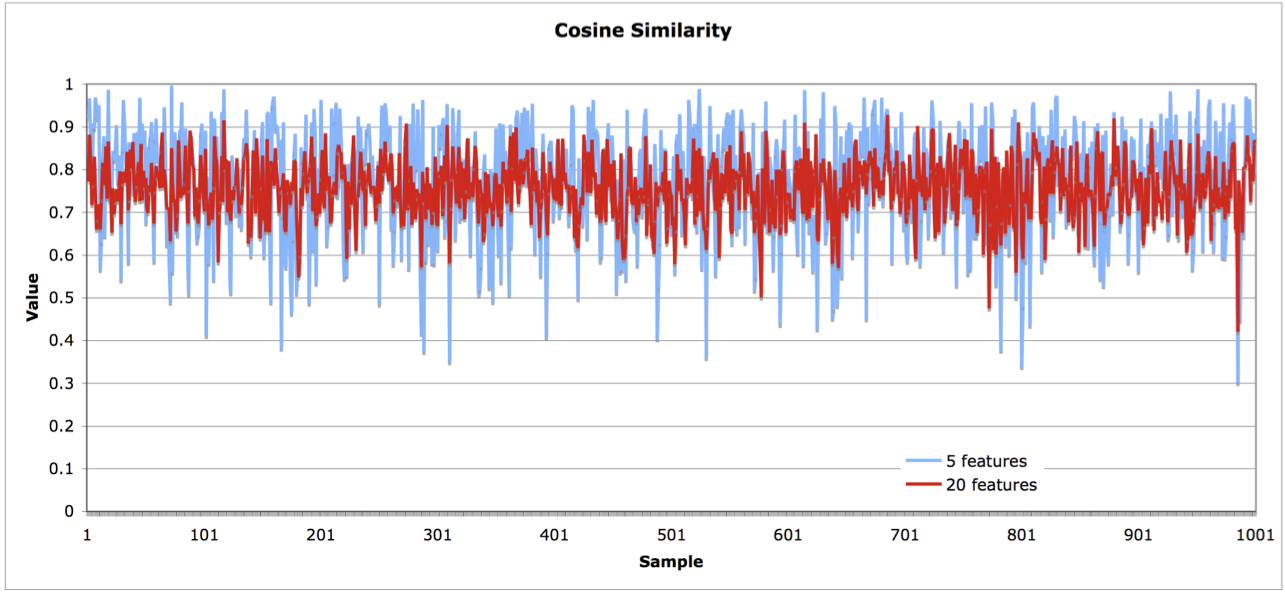}
\caption{The more dimensions used to describe objects the more similar on average things appear. This figure shows the cosine similarity between objects described by 5 and by 20 features. It is clear that in 20 dimensions similarity has a lower variance than in 5.}
\label{fig:HighDim}
\end{figure}

When the objective is to reduce the number of features used to describe data there are two strategies that can be employed. Techniques such as Principle Components Analysis (PCA) may be employed to \emph{transform} the data into a lower dimension represention. Alternatively feature selection may be employed to \emph{discard} some of the features. In using $k$-NN with high dimension data there are several reasons why it is useful to perform feature selection:
\begin{itemize}
\item[--] For many distance measures, the retrieval time increases directly with the number of features (see section \ref{sec:Complexity}).
\item[--] Noisy or irrelevant features can have the same influence on retrieval as predictive features so they will impact negatively on accuracy.
\item[--] Things look more similar on average the more features used to describe them (see Figure \ref{fig:HighDim}).
\end{itemize}

Feature Selection techniques typically incorporate a search strategy for exploring the space of feature subsets, including methods for determining a suitable starting point and generating successive candidate subsets, and an evaluation criterion to rate and compare the candidates, which serves to guide the search process. The evaluation schemes can be divided into two broad categories:
\begin{itemize}
\item[--] \textbf{Filter} approaches attempt to remove irrelevant features from the feature set prior to the application of the learning algorithm. Initially, the data is analysed to identify those dimensions that are most relevant for describing its structure. The chosen feature subset is subsequently used to train the learning algorithm. Feedback regarding an algorithm's performance is not required during the selection process, though it may be useful when attempting to gauge the effectiveness of the filter.
\item[--] \textbf{Wrapper} methods for feature selection make use of the learning algorithm itself to choose a set of relevant features. The wrapper conducts a search through the feature space, evaluating candidate feature subsets by estimating the predictive accuracy of the classifier built on that subset. The goal of the search is to find the subset that maximises this criterion.
\end{itemize}
It is worth mentioning at this point that some other classification techniques perform implicit feature selection. For instance the process of building a decision tree will very often not select all the features for use in the tree. Features not used in the tree have no role then in classification.

\textbf{Filter Techniques} Central to the Filter strategy for feature selection is the criterion used to score the predictiveness of the features. In recent years Information Gain (IG) has become perhaps the most popular criterion for feature selection. The Information Gain of a feature is a measure of the amount of information that a feature brings to the training set \cite{quinlan2014c4}. It is defined as the expected reduction in entropy caused by partitioning the training set $D$ using the feature $f$ as shown in Equation \ref{eqn:IG} where $D_v$ is that subset of the training set $D$ where feature $f$ has value $v$.
\begin{equation}\label{eqn:IG}
    IG(D,f)=Entropy(D)-\sum_{v \in values(f)}\frac{|D_v|}{|D|}Entropy(D_v)
\end{equation}
Entropy is a measure of how much randomness or impurity there is in the data set. It is defined in Equation 14 where $c$ equals the number of classes in the training set and pi is the proportion of class i in the data -- entropy is highest when the proportions are equal.
\begin{equation}
    Entropy(D)=\sum_{i=1}^c -p_i \log_2p_i
\end{equation}

In binary classification, the $Entropy(D)$can be simplified to $Entropy(D) = -p_+\log_2 p_+ - p_-\log_2 p_-$ where $p_+$ represents the class and $p_-$ the non-class.
For comparison purposes we will also consider Odds Ratio (OR)\cite{mladenic1998feature} which is an alternative filtering criterion. For binary classification OR calculates the ratio of the odds of a feature occurring in the class to the odds of the feature occurring in the non-class.

\begin{equation}
    OR(f,c)=\frac{Odds(f|c)}{Odds(f|\bar c)}
\end{equation}
Where a specific feature does not occur in a class, it can be assigned a small fixed value so that the OR can still be calculated. For feature selection, the features can be ranked according to their OR with high values indicating features that are very predictive of the class. The same can be done for the non-class to highlight features that are predictive of the non-class.

\begin{figure}[t]
\centering
\includegraphics[width=10cm]{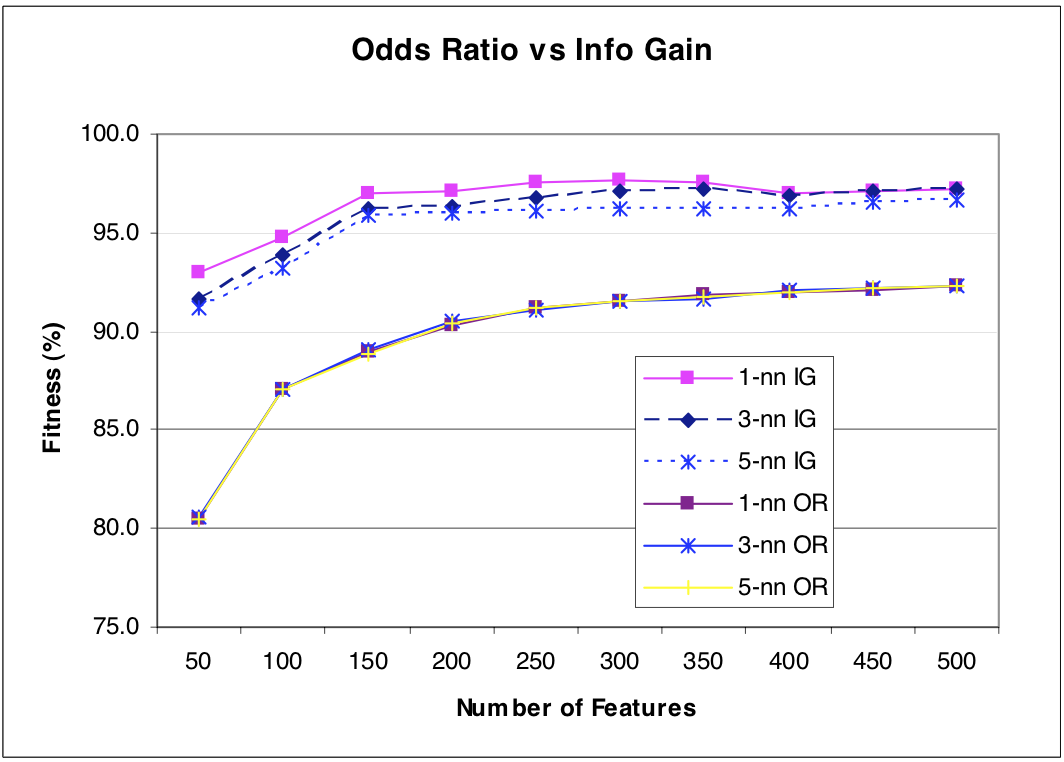}
\caption{Comparing Information Gain with Odds Ratio. Results of the average of three 10-fold cross validation experiments on a dataset of 1000 emails, 500 spam and 500 legitimate where word features only were used.}
\label{fig:IGainOR}
\end{figure}

We can look at the impact of these feature selection criteria in an email spam classification task. In this experiment we selected the $\frac{n}{2}n$ features with the
highest IG value and $n$ features each from $OR(f, spam)$ and $OR(f, nonspam)$ sets. The results, displayed in Figure \ref{fig:IGainOR}, show that IG performed significantly better than OR. The reason for this is that OR is inclined to select features that occur rarely but are very strong indicators of the class. This means that some objects (emails) are described by no features and thus have no similarity to any cases in the case base. In this experiment this occurs in 8.8\% of cases with OR compared with 0.2\% for the IG technique.
This shows a simple but effective strategy for feature selection in very high dimension data. IG can be used to rank features, then a cross validation process can be employed to identify the number of features above which classification accuracy is not improved. This evaluation suggests that the top 350 features as ranked by IG are adequate.

While this is an effective strategy for feature selection it has the drawback that features are considered in isolation so redundancies or dependancies are ignored. Two strongly correlated features may both have high IG scores but one may be redundant once the other is selected. More sophisticated filter techniques that address these issues using Mutual Information to score \emph{groups} of features have been researched by Novovi{\v{c}}ov{\'a} \emph{et al.} \cite{novovivcova2004feature} and have been shown to be more effective than these simple Filter techniques.

\textbf{Wrapper Techniques} The obvious criticism of the Filter approach to feature selection is that the filter criterion is separate from the induction algorithm used in the classifier. This is overcome in the Wrapper approach by using the performance of the classifier to guide search in feature selection -- the classifier is wrapped in the feature selection process \cite{kohavi1997wrappers}. In this way the merit of a feature subset is the generalisation accuracy it offers as estimated using cross-validation on the training data. If 10-fold cross validation is used then 10 classifiers will be built and tested for each feature subset evaluated -- so the wrapper strategy is very computationally expensive. If there are $p$ features under consideration then the search space is of size $2^p$ so it is an exponential search problem.

\begin{figure}[t]
\centering
\includegraphics[width=10cm]{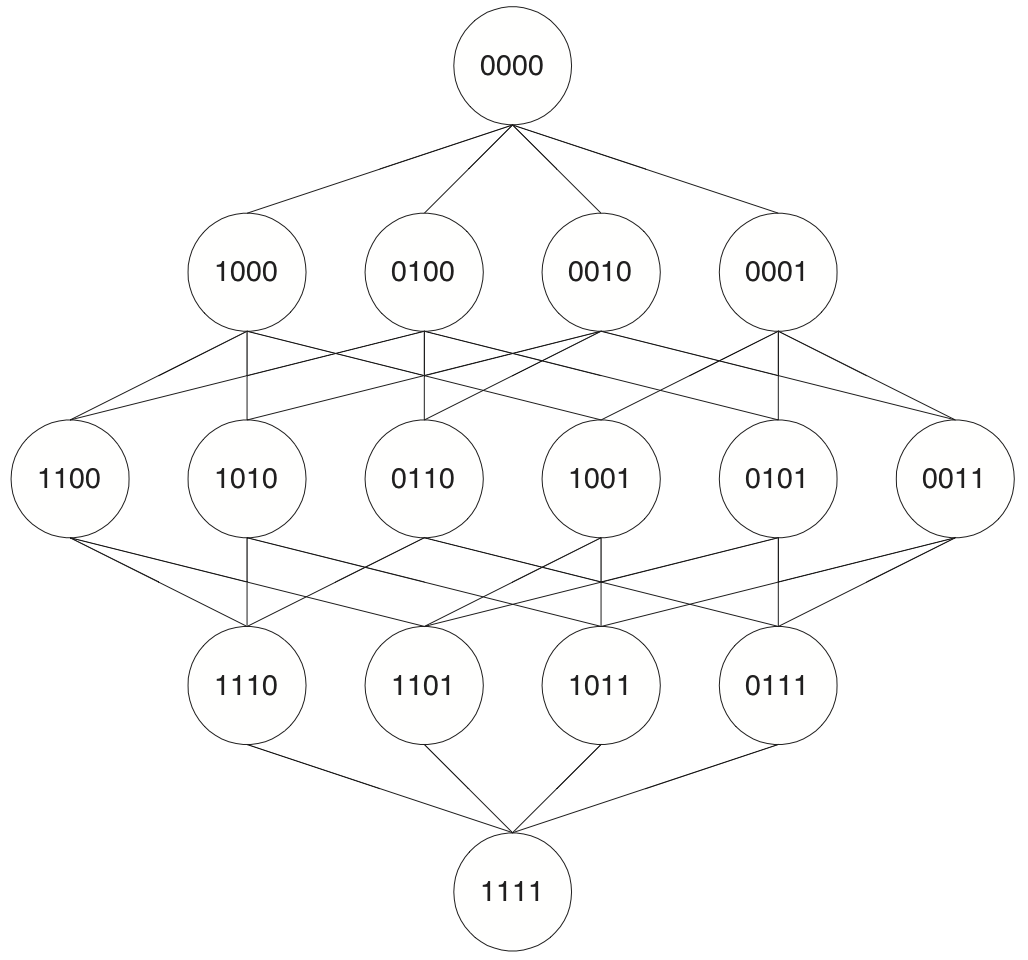}
\caption{The Feature Subspace.}
\label{fig:Wrapper}
\end{figure}

A simple example of the search space for feature selection where $p = 4$ is shown in Figure \ref{fig:Wrapper}. Each node is defined by a feature mask; the node at the top of the figure has no features selected while the node at the bottom has all features selected. For large values of $p$ an exhaustive search is not practical because of the exponential nature of the search. 
\newpage
The two most popular strategies are:
\begin{itemize}
\item[--] Forward Selection which starts with no features selected, evaluates all the options with just one feature, selects the best of these and considers the options with that feature plus one other, etc.
\item[--]Backward Elimination starts with all features selected, considers the options with one feature deleted, selects the best of these and continues to eliminate features.
\end{itemize}
These strategies will terminate when adding (or deleting) a feature will not produce an improvement in classification accuracy as assessed by cross validation.
Both of these are greedy search strategies and so are not guaranteed to discover the best feature subset. More sophisticated search strategies can be employed to better explore the search space; however, Reunanen \cite{reunanen2003overfitting} cautions that more intensive search strategies are more likely to overfit the training data.

\subsection{Instance Selection and Noise Reduction} \label{sec:NoiseRed}

The second aspect of dimension reduction is instance selection, reducing the size of the training set by removing redundant or noisy instances while maintaining or even improving performance.  This aspect of dimension reduction is explored and researched in two different areas, nearest-neighbour classification and case-based reasoning (CBR). It is known as Instance Selection or Prototype Selection by those who focus on nearest neighbour classification \cite{GarciaDCH12} and Case-base Editing or Case-base Maintenance by the CBR community \cite{mckenna2000competence}.

\begin{figure}[t]
\centering
\includegraphics[width=0.7\linewidth]{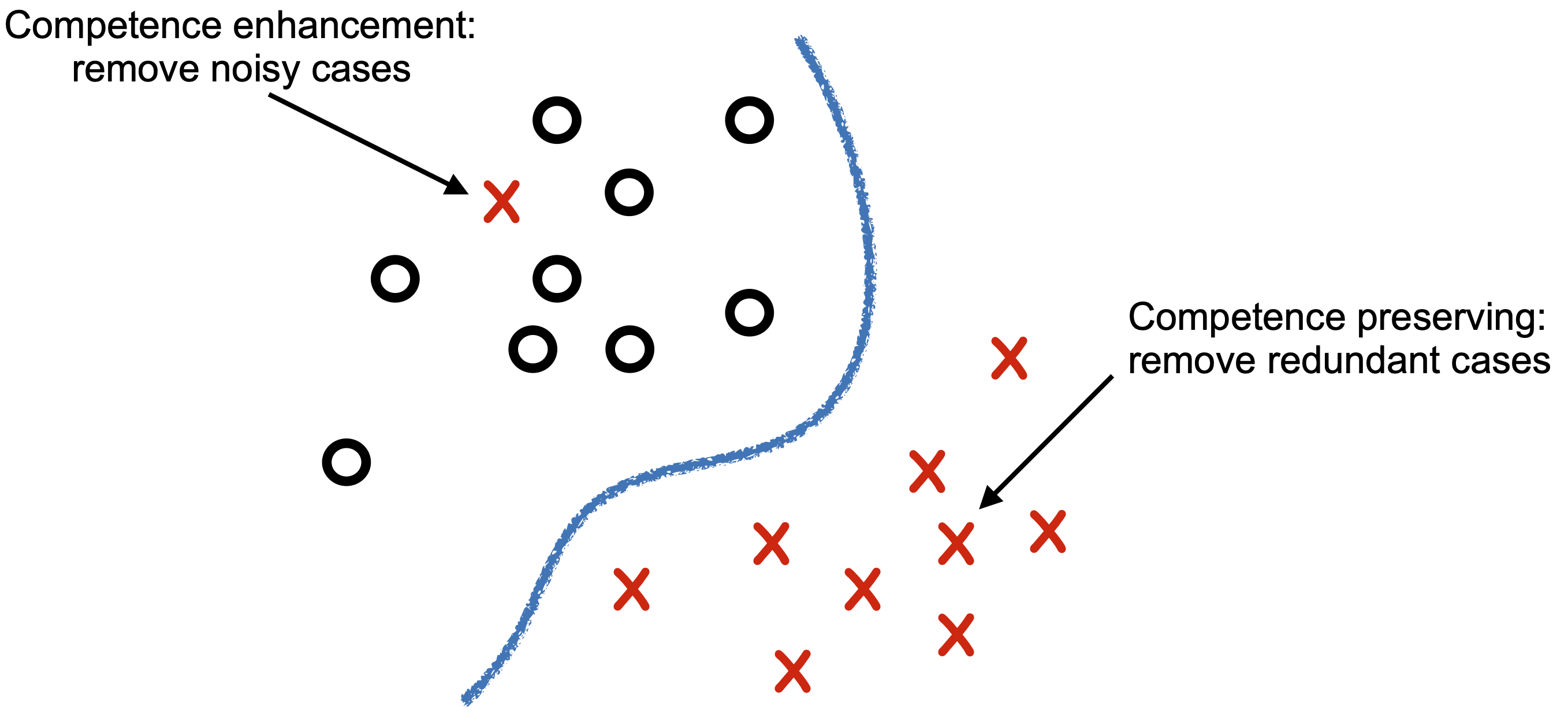}
\caption{Instance selection techniques demonstrating competence preservation (redundancy reduction) and competence enhancement (noise reduction).}
\label{fig:CBEdit}
\end{figure}

Instance selection techniques can be categorised as competence preservation or competence enhancement techniques\cite{brighton2002advances}. Competence preservation corresponds to redundancy reduction, removing superfluous instances that do not contribute to classification competence. Competence enhancement is effectively noise reduction, removing noisy or corrupt instances from the training set. Figure \ref{fig:CBEdit} illustrates both of these with a classification example, where instances of one class are represented by stars and instances of the other class are represented by circles. Competence preservation techniques aim to remove internal instances in a cluster of instances  of the same class and can predispose towards preserving noisy instances as exceptions or border cases.

Noise reduction on the other hand aims to remove noisy or corrupt instances but can remove exceptional or border instances which may not be distinguishable from true noise, so a balance of both can be useful.  Techniques which combine a balance of both redundancy reduction and noise removal are known as hybrid approaches. 

Editing strategies normally operate in one of two ways; \emph{incremental} which involves adding selected instances from the training set to an initially empty edited set, and \emph{decremental} which involves contracting the training set by removing selected instances.

\textbf{Early Techniques} 
An early competence preservation technique is Hart’s Condensed Nearest Neighbour (CNN) \cite{hart1968condensed}. CNN is an incremental technique which adds to an initially empty edited set any instance from the training set that cannot be classified correctly by the edited set. This technique is very sensitive to noise and to the order of presentation of the training set instances, in fact CNN by definition will tend to preserve noisy instances. 
Improvements on CNN included the Selective NN (SNN)  \cite{ritter1975algorithm} which imposes the rule that every instance in the training set must be closer to an instance of the same class in the edited set than to any other training instance of a different class. The Reduced NN Rule\cite{gates1972reduced} took the opposite, decremental, approach removing a instance from the training set where its removal does not cause any other instance to be misclassified. This technique will allow for the removal of noisy cases but is sensitive to the order of presentation of cases.

Competence enhancement or noise reduction techniques start with Wilson’s Edited Nearest Neighbour (ENN) algorithm \cite{wilson1972asymptotic}, a decremental strategy, which removes instances from the training set which do not agree with their $k$ nearest neighbours. These instances are considered to be noise and appear as exceptional examples in a group of instances of the same class. Extentions to ENN include the repeated ENN (RENN) and the \emph{all k-NN} algorithms\cite{tomek1976experiment}. Both make multiple passes over the training set, the former repeating the ENN algorithm until no further eliminations can be made from the training set and the latter using incrementing values of $k$. These techniques focus on noisy or exceptional instances and do not result in the same storage reduction gains as the competence preservation approaches.

 Hybrid techniques were introduced with a series of instance based learning IB\emph{n} algorithms\cite{aha1991instance}. IB2 is similar to CNN adding only instances that cannot be classified correctly by the reduced training set. IB2’s susceptibility to noise is handled by IB3 which records how well instances are classifying and only keeps those that classify correctly to a statistically significant degree. Other researchers have provided variations on the IB$n$ algorithms \cite{brodley1993addressing,cameron1992minimum,zhang1992selecting}. 

\textbf{Competence-Based Case-Base Editing} Approaches to case-base editing build a competence model of the training data and use the competence properties of the cases to determine which cases to include in the edited set. Measuring and using case competence to guide case-base maintenance was first introduced by Smyth and Keane \cite{smyth1995remembering} and developed by Zhu and Yang \cite{zhu1999remembering}. Smyth and Keane \cite{smyth1995remembering} introduce two important competence properties, the \emph{reachability} and \emph{coverage} sets for a case in a case-base. The \emph{reachability set} of a case $c$ is the set of all cases that can successfully classify $c$, and the coverage set of a case $c$ is the set of all cases that $c$ can successfully classify. The coverage and reachability sets represent the local competence characteristics of a case and are used as the basis of a number of editing techniques.

A family of competence-guided editing methods for case-bases combine both incremental and decremental strategies using a combination of rules \cite{mckenna2000competence}: 
\begin{enumerate}[label=(\roman*)]
\item an \emph{ordering policy} for the presentation of the cases that is based on the competence characteristics of the cases,
\item an \emph{addition rule} to determine the cases to be added to the edited set,
\item a \emph{deletion rule} to determine the cases to be removed from the training set
and
\item an \emph{update policy} which indicates whether the competence model is updated
after each editing step.
\end{enumerate}

One of these algorithms, Conservative Redundancy Removal (CRR) \cite{delany2004analysis} is included in the assessment in section \ref{sec:editeval}.  This algorithm is similar in concept to the FCNN rule \cite {Angiulli07} which can be applied to huge collections of data.  

Other approaches also use the coverage and reachability properties of cases.  Iterative Case Filtering (ICF) \cite{brighton2002advances} is a decremental strategy contracting the training set by removing those cases $c$, where the number of other cases that can correctly classify $c$ is higher that the number of cases that $c$ can correctly classify. This strategy focuses on removing cases far from class borders. After each pass over the training set, the competence model is updated and the process repeated until no more cases can be removed. ICF includes a pre-processing noise reduction stage, effectively RENN, to remove noisy cases. McKenna and Smyth compared their family of algorithms to ICF and concluded that the overall best algorithm of the family delivered improved accuracy (albeit marginal, 0.22\%) with less than 50\% of the cases needed by the ICF edited set \cite{mckenna2000competence}.

Wilson and Martinez \cite{WilsonM00} present a series of reduction techniques called DROP1 to DROP5\footnote{Three of these algorithms were originally published in \cite{wilson1997instance} as Reduction Techniques (RT1 to RT3).}  which, although published before the definitions of coverage and reachability, could also be considered to use a competence model. 
They define the set of associates of a case $c$ which is comparable to the coverage set of McKenna and Smyth except that the associates set will include cases of a different class from case $c$ whereas the coverage set will only include cases of the same class as $c$. 

The DROP$n$ algorithms use a decremental strategy. 
%DROP1, the basic algorithm, removes a case $c$ if at least as many of its associates would still be classified correctly without $c$. This algorithm focuses on removing noisy cases and cases at the centre of clusters of cases of the same class as their associates which will most probably still be classified correctly without them. DROP2 fixes the order of presentation of cases as those furthest from their nearest unlike neighbour (i.e. nearest case of a different class) to remove cases furthest from the class borders first. DROP2 also uses the original set of associates when making the deletion decision, which effectively means that the associate’s competence model is not rebuilt after each editing step which is done in DROP1. DROP3 adds a noise reduction pre-processing pass based on Wilson’s noise reduction algorithm. The noise reduction algorithm is adapted slightly in DROP4 to be more selective and only removes cases misclassified by their neighbours if their removal does not impact the classification of other instances.  DROP5 adapts DROP2 by including one initial noise reduction pass which uses the reverse of the DROP2 ordering, i.e. instances that are nearest to their nearest unlike neighbour are considered first. This has the effect of smoothing the decision boundary.  
Their comprehensive evaluation found DROP3 to be the best mix of generalisation accuracy and  storage requirements, performing consistently well in comparison with other reduction techniques. 
A comparison of  ICF against DROP3 found that neither algorithm consistently out performed the other and both represented the “cutting edge in instance set reduction techniques” \cite{brighton2002advances}.

\begin{figure}[t]
\begin{multicols}{3}
    \includegraphics[width=0.98\linewidth]{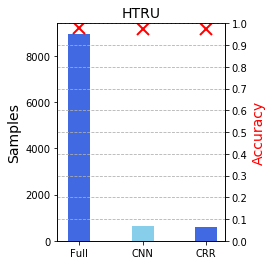}
     \par 
    \includegraphics[width=0.95\linewidth]{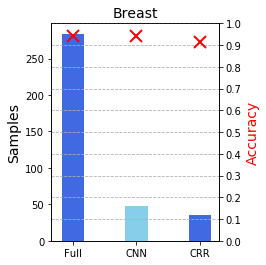}
     \par 
    \includegraphics[width=\linewidth]{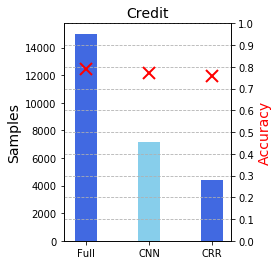}\par 
\end{multicols}
\caption{The impact of CNN and CRR on training set size and accuracy.}
\label{fig:edit}%
\end{figure}

\subsubsection{Instance Selection Performance}\label{sec:editeval}
Figure \ref{fig:edit} shows the impact of two instance selection techniques (CNN \& CRR) on training set size and generalisation accuracy. The evaluation shows that, at least for some datasets, the training set size can be dramatically reduced with almost no impact on generalisation accuracy. If there is a lot of redundancy in the training data, dramatic speedup can be achieved through instance selection without any significant impact on accuracy. 

\section{Conclusion: Advantages and Disadvantages}
$k$-NN is very simple to understand and easy to implement. So it should be considered in seeking a solution to any classification problem. Some advantages of $k$-NN are as follows (many of these derive from its simplicity and interpretability):
\begin{itemize}
\item[--] Because the process is transparent, it is easy to implement and debug.
\item[--] $k$-NN can be applied to data that cannot be described as a feature vector provided a similarity measure is available. Thus $k$-NN can be used in situations where other ML mechanisms will not be applicable. 
\item[--] In situations where an explanation of the output of the classifier is useful, $k$-NN can be very effective if an analysis of the neighbours is useful as explanation.
\item[--] There are some noise reduction techniques that work only for $k$-NN that can
be effective in improving the accuracy of the classifier \cite{delany2004analysis}.
\item[--] In some circumstances, speed-up mechanisms such as Kd-Trees or Ball Trees can improve retrieval times without any loss of accuracy.
\item[--] Approximate Nearest Neighbour techniques can greatly improve retrieval times, sometimes with minimal impact on accuracy \cite{kleinberg1997two}. 
\end{itemize}

These advantages of $k$-NN, particularly those that derive from its interpretability, should not be underestimated. On the other hand, some significant disadvantages are as follows:
\begin{itemize}
\item[--] Because all the work is done at run-time, $k$-NN can have poor run-time performance if the training set is large.
\item[--] $k$-NN is very sensitive to irrelevant or redundant features because all features contribute to the similarity (see Eq. \ref{eqn:dist}) and thus to the classification. This can be ameliorated by careful feature selection or feature weighting.
\item[--] On very difficult classification tasks, $k$-NN may be outperformed by more \emph{exotic} techniques such as Support Vector Machines or Neural Networks.
\end{itemize}

\appendix
\section{Appendix I: Python Code}\label{app:code}

The GitHub repository\footnote{\url{https://github.com/PadraigC/kNNTutorial}} associated with this paper contains the following Python Notebooks:
\begin{itemize}
    \item \textsf{kNN-Basic}: Code for a basic $k$-NN classifier in \textsf{scikit-learn}.
    \item \textsf{kNN-Correlation}: How to use correlation as the $k$-NN metric in \textsf{scikit-learn} (see section \ref{sec:corr}).
    \item \textsf{kNN-Cosine}: How to use Cosine as the $k$-NN metric in \textsf{scikit-learn}. Using Cosine similarity for text classification (section \ref{sec:cos}). 
    \item \textsf{kNN-DTW}: Using the \textsf{tslearn} library for time-series classification using DTW (section \ref{sec:DTW}).
    \item \textsf{kNN-Speedup}: \textsf{scikit-learn} provides some options for speeding up nearest neighbour retrieval. This notebook tests the speedup on four datasets (section \ref{sec:speedup}). 
    \item \textsf{kNN-Annoy}: Testing the speedup offered by the Approximate Nearest Neighbour implementation in \textsf{annoy} (section \ref{sec:speedup}).
    \item \textsf{kNN-PCA}: Some code to use PCA to estimate the intrinsic dimension of the four datasets. 
    \item \textsf{kNN-InstSel}: An evaluation of the impact of two Instance selection algorithms (CRR \& CNN) on training set size and accuracy. 
\end{itemize}

% \section{Stuff to Add?}
% \vbox{% prevent page break in list
% \begin{itemize}
%     \item k-NN Regression
%     \item SpeedUP - really rewrite + add
%     \item time-series 1-NN
%     \item Divide measures into time-series and multimedia?
%     \item Divide speedup into algs and editing
%     \item kNN in sklearn
% \end{itemize}
% }

% Introduction
% Similarity Measures 
 % Metrics
 % Correlation
 % Multimedia
 % Time Series
% Speed-up
        % - mention intrinsic dimensionality
 
\bibliographystyle{plain}  
\bibliography{kNN}  %%% Remove comment to use the external .bib file (using bibtex).

\end{document}